# CurateGPT: A flexible language-model assisted biocuration tool

J Harry Caufield[1,*], Carlo Kroll[2], Shawn T O'Neil[3], Justin T Reese[1], Marcin P Joachimiak[1], Harshad Hegde[1], Nomi L Harris[1], Madan Krishnamurthy[3], James A McLaughlin[4], Damian Smedley[2], Melissa A Haendel[3], Peter N Robinson[5], Christopher J Mungall[1]

[1]Lawrence Berkeley National Laboratory, Berkeley, CA, USA
[2]Queen Mary University of London, London, UK
[3]University of North Carolina at Chapel Hill, Chapel Hill, NC, USA
[4]European Bioinformatics Institute (EMBL-EBI), Hinxton, UK
[5]Berlin Institute of Health - Charité, Universitätsmedizin, Berlin, Germany

* Corresponding author <jhc@lbl.gov>

## Abstract

Effective data-driven biomedical discovery requires data curation: a time-consuming process of finding, organizing, distilling, integrating, interpreting, annotating, and validating diverse information into a structured form suitable for databases and knowledge bases. Accurate and efficient curation of these digital assets is critical to ensuring that they are FAIR, trustworthy, and sustainable. Unfortunately, expert curators face significant time and resource constraints. The rapid pace of new information being published daily is exceeding their capacity for curation.

Generative AI, exemplified by instruction-tuned large language models (LLMs), has opened up new possibilities for assisting human-driven curation. The design philosophy of *agents* combines the emerging abilities of generative AI with more precise methods. A curator's tasks can be aided by agents for performing reasoning, searching ontologies, and integrating knowledge across external sources, all efforts otherwise requiring extensive manual effort.

Our LLM-driven annotation tool, CurateGPT, melds the power of generative AI together with trusted knowledge bases and literature sources. CurateGPT streamlines the curation process, enhancing collaboration and efficiency in common workflows. Compared to direct interaction with an LLM, CurateGPT's agents enable access to information beyond that in the LLM's training data and they provide direct links to the data supporting each claim. This helps curators, researchers, and engineers scale up curation efforts to keep pace with the ever-increasing volume of scientific data.

# Introduction

Databases and knowledge bases require high-quality curation to ensure integrity, validity and FAIRness. Curation is the task of finding and contextualizing diverse information into a structured form suitable for populating these knowledge resources. Curation is a time-consuming, painstaking process, so there is interest in applying machine learning (ML) and natural language processing (NLP) techniques to assist with the labor. However, the efficacy of these approaches is limited by the unavailability of domain-specific training data and an inability to incorporate domain knowledge (1). ML approaches have yet to be integrated into production curation systems, save for PubTator's biomedical literature annotation (2) and for certain circumscribed tasks such as literature triage (3–5). These automated approaches can help to address the ever growing volume of scientific data and publications: PubMed alone adds roughly one million new citations to its index each year (as per the MEDLINE Citation Counts as of January 2023; https://www.nlm.nih.gov/bsd/medline_cit_counts_yr_pub.html).

Dependability is crucial for curated data resources, so ML-based curation support approaches without demonstrable reliability are of limited value. Curation workflows based around ML-driven tools pose definite risks: they may miss relevant details, introduce factual inaccuracies, or even generate fully unsupported data relationships, all without sufficient provenance details to identify the source of error. To identify these issues, curators must sift through the automated predictions – often a more time-intensive task than curating directly from source. The current consensus is that, on their own, traditional NLP approaches are not a good substitute for manual curation, particularly in the biomedical domain (6,7). NLP methods have a substantial history of assisting with curation (8–10). The emergence of large language models (LLMs), such as GPT-4, Claude, Gemini, and Llama3, has introduced potential solutions for the limitations of NLP. LLMs have highly general information processing capabilities and do not require retraining for each task or use case. The advantages LLMs offer over more traditional NLP approaches, including their flexibility and rapid adaptability to varied tasks (11), may render them more feasible for literature curation applications. Researchers developing new sets of annotated biomedical text have found LLMs to be particularly efficient (12), as have medical informatics researchers in applying the models to clinical decision support (13). Even so, evaluating LLM output continues to be a job best performed by human curators.

Biomedical curation is often performed using customized user interfaces with few avenues for integrating AI assistance. We hypothesized that the creation of an AI-powered integrated development environment (IDE) could transform the practice of biocuration. We have developed the prototype of such an IDE, called CurateGPT (https://github.com/monarch-initiative/curate-gpt). This system supports curation of an extensive variety of data types, ranging from tabular data to nested schemas common to biomedical knowledge bases (e.g., UniProt (14) or the Alliance of Genome Resources (15)) and ontologies (e.g., Gene Ontology (16) or Cell Ontology (17)). CurateGPT is based around a flexible document store and vector database with dynamic access to online resources, including Wikipedia, PubMed, and other NCBI databases. It integrates several generative AI engineering techniques: Retrieval Augmented Generation (RAG), which allows for the most relevant documents to be used as context for LLM queries, obviating the need for specialized training; methods to extract structured information from unstructured text, including our own SPIRES algorithm (18); and new approaches for retrieving evidence from existing statements.

CurateGPT supports multiple steps of the curation process, from schema design to information retrieval, structuring data, and finding supporting evidence. It incorporates the LinkML data modeling framework to manage schemas for collections of curated assertions (19). Its incorporation of RAG methods allows it to produce results adherent to extant knowledge bases; such an approach was recently found to be very effective in curating biosample metadata (20). Here we provide an overview of the CurateGPT architecture, capabilities of its individual agents, and demonstrations of its workflows.

# Methods

The CurateGPT architecture centers around a collection of independent agents that can be invoked in three ways: via APIs; from the command line; and via a prototype graphical interface. Each agent is the combination of a function and an LLM (or more specifically, the interface to an LLM). These agents are described below. Further technical detail is provided in the System Architecture section.

## CurateGPT Agents

CurateGPT provides agents to enable the following tasks:
- `Search` - find entries within an ontology or document collection relating to a text query.
- `Chat` - query a knowledge base with natural language.
- `Curate` - generate a new ontology class or an object fitting a provided data model.
- `Extract` - transform unstructured text into structured knowledge.
- `CiteSeek` - find citations for a claim, based on the contents of a knowledge base.
- `Match` - identify potential mappings for a term within an ontology.
- `Bootstrap` - create a seed schema and knowledge base from a description.

Combinations of these agents can power a curation workflow. For example, the process of isolating specific claims from new literature may start with the `Curate` command and then use the `CiteSeek` command to obtain additional sources for context.

All CurateGPT interfaces offer a variety of settings, including:
- Which LLM to use. CurateGPT agents are largely model architecture-agnostic and will work with a variety of LLM interfaces and local implementations, including open models such as Llama3 (21) and its successors.
- Which collection of input knowledge to operate on and source examples from. This may be an ontology, a controlled vocabulary, or some other structured data resource.
- What collection to use for background sources, such as PubMed or Wikipedia.

In addition to providing visual options for selecting the above options, the prototype graphical interface for CurateGPT can be deployed on a local system or adapted into a remotely-accessible toolset. It is built on the open-source Streamlit (https://streamlit.io/) user interface framework to encourage rapid implementation and re-use. Examples for each agent in the sections below are described as they appear in the graphical interface; all example outputs are generated with the OpenAI GPT-4o model.

## Search

CurateGPT's `Search` agent allows curators to find values within their data matching a text query. Search results can be displayed as a table or using a reduced dimensionality visualization such as PCA, t-SNE, or UMAP (in the graphical interface, this is titled Cluster Search). Though search is not the primary function of CurateGPT, search underpins many operations in the curation process. Either pre-loaded static collections can be searched, or dynamic search can be performed on wrapped remote resources accessed via API, such as PubMed or Wikipedia. The results of searches can be selected to be placed in a cart, for use either as (a) curation/refinement or (b) background knowledge in other curation tasks.

## Chat

The CurateGPT `Chat` agent allows curators to ask natural language questions about one or more data collections. The `Chat` agent will find the most relevant objects relating to the question (specifically, those with the least difference between embeddings of the query and the collection data), then include these as background context. The model will be prompted to cite these objects in the response. The response is then parsed, and matched back to the original input objects. The answers are provided with inline citations, hyperlinked to specific sources further down on the page (see example in **Table 1**). Compared to direct interaction with an LLM through its chat interface, the `Chat` agent provides two major advantages: (a) it derives information from custom input documents, including documents not present in the LLM training data; and (b) it provides identifiers for relevant data, giving the user additional opportunities to verify the LLM response. As part of the chat prompt, CurateGPT provides the LLM with relevant information from the custom input, so answers will pertain to those additional documents. Further technical details are discussed in the System Architecture section below.

**Table 1**. Example input and output of the `Chat` agent. Outputs are truncated for brevity. A) The agent is instructed to use the Food Ontology as the knowledge collection. B) The agent is instructed to use Wikipedia as the knowledge collection, through its API. The provided title refers to a Wikipedia page (https://en.wikipedia.org/wiki/Breakfast_cereal).

| **A** | |
|---|---|
| Input | `What food products may contain dairy ingredients and gluten?` |
| Output summary | `Based on the provided references, the following food products may contain both dairy ingredients and gluten:` |
| | Cake Food Product: This product can include dairy ingredients such as butter or milk and may also contain gluten from flour, which is a common ingredient in cakes 3.<br>... |
| Output reference | Reference 3<br>`id`: CakeFoodProduct<br>`label`: cake food product<br>`definition`: A food that is usually sweet and often baked...<br>`relationships`:<br>- `predicate`: subClassOf<br>  `target`: BakeryFoodProduct<br>`original_id`: FOODON:00001278 |
| **B** | |
| Input | What food products may contain dairy ingredients and gluten? |
| Output summary | Food products that may contain both dairy ingredients and gluten include:<br>Various processed and packaged foods often contain both gluten and dairy due to their formulation and the use of these ingredients as additives. Some examples include baked goods such as cakes, cookies, and pastries, which typically use wheat flour (containing gluten) and may also have dairy products like butter and milk.<br>Breakfast cereals are another category where both dairy and gluten can be commonly found. These cereals are usually made from processed cereal grains, many of which contain gluten, and are often served with dairy products like milk 7.<br>... |
| Output reference | Reference 7<br>`title`: Breakfast cereal<br>`extract`: "Breakfast cereal is a category of food, including food products... |

## Curate

The `Curate` agent supports CurateGPT's ability to generate new ontology entries and structured data elements directly from text and the parameterized knowledge within an LLM. The `Curate` agent generates an object to add to a collection given some quantity of seed information. The seed information is provided as separate, distinct values for a single entry, but may be quite incomplete: a single label or description alone is enough to start building a more extensively described entry. In an ontology such as the Human Phenotype Ontology (HPO) (22), for example, providing the `Curate` agent with a class label not already in the ontology will yield a new class object that includes an identifier in the expected format, a definition similar in structure to others in the ontology, and relationships connecting the new class to others.

Optionally, a curator may use the LLM to generate a background description of the provided information along with additional details and context. The newly generated object may then be added directly to the local collection. If the agent is provided with the label for a new entry in HPO or the Environment Ontology (ENVO; (23)), for example, it will return an object such as those shown in **Table 2**.

**Table 2**. Example input and output of the `Curate` agent. A) The agent is instructed to generate a new class for the Ontology for Biomedical Investigations. B) The agent is instructed to generate a new class for the Environment Ontology.

| A | |
|---|---|
| Input | `label: Fingernail specimen` |
| Output | `label: Fingernail specimen`<br>`id: FingernailSpecimen`<br>`definition: A specimen that is derived from a fingernail.`<br>`relationships:`<br>`- predicate: subClassOf`<br>`  target: SpecimenFromOrganism` |
| **B** | |
| Input | `label: suburban stormwater` |
| Output | `label: suburban stormwater`<br>`id: SuburbanStormwater`<br>`definition: Stormwater which accumulates in a suburban ecosystem.`<br>`relationships:`<br>`- predicate: LocatedIn`<br>`  target: SuburbanBiome`<br>`- predicate: LocatedIn`<br>`  target: AreaOfResidentialDevelopment`<br>`- predicate: subClassOf`<br>`  target: Stormwater` |

## Extract

The `Extract` agent provides example-based extraction (also known as structured object autocomplete) from raw text. Unlike the `Curate` agent, which creates a new ontology object given an incomplete set of initial values, the `Extract` agent requires no separation of the input text into distinct components. The input is parsed directly into a structured form corresponding with the target collection, with fields populated using background knowledge wherever possible. Additional rules may also be passed to the LLM to guide generation of the extracted object. An example is shown in **Table 3**.

**Table 3**. Example inputs and outputs of the `Extract` agent. A) The agent is instructed to generate a new class for the Human Phenotype Ontology based on extraction from the input text (a selection from the StatPearls article on Intercostal Neuralgia). B) The agent is instructed to generate a new class for the Medical Action Ontology based on extraction from the input text.

| A | |
|---|---|
| Input | `Intercostal neuralgia is characterized by neuropathic pain in the distribution of affected intercostal nerve(s) (along the ribs, chest, or abdomen) that commonly manifests as a sharp, aching, radiating, burning, or stabbing pain and may be associated with paresthesia such as numbness and tingling. The pain may be intermittent or constant and typically presents either as a band-like pain wrapping along the chest and back or in a thoracic dermatomal pattern. Pain may last for a prolonged period and may continue long after the inflicting disease process has subsided.` |
| Additional Instructions | `Limit the definition to one sentence and 20 words or less.` |
| Output | `id: IntercostalNeuralgia`<br>`label: Intercostal neuralgia`<br>`definition: Intercostal neuralgia is characterized by neuropathic pain along the ribs, chest, or abdomen.`<br>`relationships:`<br>`- predicate: subClassOf`<br>`  target: Neuralgia` |
| **B** | |
| Input | `An endobronchial valve system may be implanted in the airways of the lungs.` |
| Output | `id: EndobronchialValveSystemImplantation`<br>`label: endobronchial valve system implantation`<br>`definition: An endobronchial valve system may be implanted in the airways of the lungs.`<br>`relationships:`<br>`- predicate: subClassOf`<br>`  target: MedicalDeviceImplantation` |

As with other agents, better results may be achieved by allowing CurateGPT to access more detailed information for use in RAG. This may include PubMed, Wikipedia, and local document collections. Additionally, CurateGPT can directly parse the content of a PubMed entry or a PubMed Central full text.

CurateGPT supports three strategies for extraction of structured objects: Basic Extraction, OpenAI Functions, and Recursive Extraction. These are a mixture of schema-free and schema-based. The basic extraction technique is schema-free, and relies on exemplar objects sourced from the input collection (i.e., those provided by the user). The LLM then generalizes based on these examples. As such, it is designed to be used in conjunction with RAG techniques. Up to 20 of the most similar objects are provided as exemplars and the LLM is prompted to generate a new object conforming to the same structure in response to the question. The OpenAI functions approach is specific to OpenAI models, and requires a schema in either Pydantic, JSON-Schema, or LinkML. The recursive extraction approach uses the method described in (18), SPIRES, but without the ontology matching step.

## CiteSeek

The `CiteSeek` agent is a central component of CurateGPT. It supports an essential task in curation: retrieving citations providing evidence for assertions. Identifying supporting literature is of particular value to building and curating relationships within ontologies. The `CiteSeek` agent is particularly powerful because it retrieves supporting information from external sources including PubMed and Wikipedia rather than the LLM alone (24,25). On their own, LLMs may frequently provide incomplete support for provided claims. The resulting citations are therefore much less likely to be incomplete or purely generated results and are much more likely to correspond to extant publications. The agent provides human-readable context for each, limiting the time curators must spend on hunting for pertinent text in each reference.

Retrieving citations with the `CiteSeek` agent works well when retrieving records through the PubMed API. For example, if we use `CiteSeek` with the PubMed API (see **Table 4A**), the agent first uses the LLM to identify specific search terms, then constructs a query to retrieve relevant literature records through the API (this is a crucial preprocessing step; without it, the example shown in the table yields no results). It then assembles a prompt instructing the LLM to examine the records and produce structured representations of their support. This may include instances where a statement from the literature supports, partially supports, refutes, finds no evidence for, or disagrees with the details of a statement. Alternatively, the input may be provided as a structured relationship (**Table 4**). When complete, the agent returns a textual summary describing each citation along with bibliographic details for the source supporting this claim.

**Table 4**. Example inputs and outputs of the `CiteSeek` agent. For all examples, the table includes a selection of the textual reference summary and the first cited reference, with truncated abstract text. A) Input and output for citations retrieved through the PubMed API. B) Input and output for citations within the Medical Actions Ontology (MAXO).

| A | | |
|---|---|---|
| Input | ```Premature graying of hair can be caused by genetics, nutritional deficiencies, medical conditions, stress, smoking, and environmental factors``` | ```subject: Ramucirumab```<br>```predicate: treats```<br>```object: gastric cancer``` |

| Output summary | The references provided contain several pieces of evidence that support the assertion that premature graying of hair can be caused by genetics, nutritional deficiencies, medical conditions, stress, smoking, and environmental factors:<br><br>Genetics: Reference 1 highlights the strong association between family history of premature hair graying (PHG) and its occurrence. PHG is significantly associated with a history of PHG in family members, such as the mother, father, maternal grandparent, paternal grandparent, and siblings... | The references provided offer substantial evidence supporting the statement that "Ramucirumab treats gastric cancer."<br><br>Reference 3 discusses the use of ramucirumab as a treatment for gastric cancers. It highlights that ramucirumab is a recombinant monoclonal antibody targeting VEGFR2, showing second-line effectiveness for patients with gastric carcinomas... |
|---|---|---|
| Output reference | id: PMID:31523106<br>title: Evaluation of Physiological, Psychological, and Lifestyle Factors Associated with Premature Hair Graying.<br>abstract: 'BACKGROUND: Canities, or hair graying, is believed to be...<br>pmcid: PMCID:PMC6706993 | id: PMID:27149032<br>title: Ramucirumab for the treatment of gastric cancers, colorectal adenocarcinomas, and other gastrointestinal malignancies.<br>abstract: 'INTRODUCTION: The use of antiangiogenic strategy... |
| **B** | | |
| Input | subject: chorionic villus sampling<br>predicate: diagnoses<br>object: fetal genetic disease | |
| Output summary | Based on the background facts provided, chorionic villus sampling (CVS) is defined as a prenatal test where a sample of chorionic villi from the placenta is removed for testing. It is a type of prenatal genetic testing, which aligns with the objective of diagnosing fetal genetic diseases 1. CVS allows for genetic analysis and can provide information about genetic disorders in the developing fetus. Therefore, the evidence supports that chorionic villus sampling is used in the diagnosis of fetal genetic diseases. | |
| Output reference | id: ChorionicVillusSampling<br>label: chorionic villus sampling<br>definition: A prenatal test in which a sample of chorionic villi is removed from the placenta for testing. The sample can be taken | |

```
            through the cervix (transcervical) or the abdominal wall
            (transabdominal). Mayo Clinic
          relationships:
          - predicate: subClassOf
            target: PrenatalGeneticTesting
          - predicate: subClassOf
            target: ClinicalBiopsyBySite
          - predicate: subClassOf
            target: WholeBodyExamination
          - predicate: subClassOf
            target: TherapeuticProcedureOfOrganismSubstance
          - predicate: subClassOf
            target: TherapeuticProcedureOfOrgan
          - predicate: subClassOf
            target: DiagnosticProcedureOfBodySubstance
          - predicate: subClassOf
            target: DiagnosticProcedureOfOrgan
          original_id: MAXO:0000536
```

The `CiteSeek` agent also supports using Ontology collections as citation sources (**Table 4B**). As with a literature search, in this case CiteSeek finds one or more records supporting the claim. The generated summary includes further justification for supporting the claim based on the record's placement in the ontology.

## Match

Mapping terms between ontologies and controlled vocabularies is a frequently necessary task when curating and applying these resources. It is very time-consuming for curators, as they must link identifiers that denote entities or concepts with varying degrees of abstraction and similarity. We have previously outlined some of these challenges in our paper on Simple Standards for Sharing Ontology Mappings (SSSOM) (26). CurateGPT supports mapping curation through the `Match` agent, which takes the label of a concept to match as input, queries the data collection for the best matches by vector distance, then queries the LLM regarding the best options out of the set of potential matches. For example, searching for the best matches for the text "round red fruit with many seeds in it" across the Food Ontology (FOODON) yields ten results by default, including "sweet red bell pepper" (FOODON:00003485), "red currant" (FOODON:00003766), and "red raspberry" (FOODON:00003729). The `Match` agent selects "red raspberry" as the best match. This works well for cross-lingual matches as well: instructing the `Match` agent to search for matches for the Polish word "wątroba" in OBI yields the class for the corresponding concept with its English label, "liver" (imported from the Uberon anatomy ontology, UBERON:0002107).

## Bootstrap

The `Bootstrap` agent facilitates generation of the initial version of a knowledge base. Starting with a brief description, the agent uses the LLM to produce a corresponding LinkML schema.

The schema (or, if preferred, an entirely different schema) may then be used to generate a data sample. This approach encourages the LLM to provide additional classes, attributes, and enumerations beyond those initially provided by the user, so the resulting schema can provide a comprehensive foundation for a new knowledge collection. For example, the `Bootstrap` agent may be provided an initial configuration as shown in **Table 5**. It will then generate a full schema (see **Supplementary Table 1**) and corresponding data (see **Supplementary Table 2**).

**Table 5**. Example input for the `Bootstrap` agent. The agent generates a schema and data following this schema (limited to a single entry here for brevity).

| Input | ```
kb_name: livestock_antibiotics
description: A knowledge base of antibiotics used in raising livestock
attributes: associated_animals, common_name, common_uses, side_effects
main_class: Antibiotic
``` |
|---|---|
| Output data | ```
name: Tetracycline
description: Tetracycline is a broad-spectrum antibiotic used in veterinary medicine.
associated_species:
  - species_name: Cattle
    notes: Often used to treat respiratory tract infections.
  - species_name: Swine
    notes: Commonly used to manage enteric diseases.
common_uses:
  - use_description: Treating bacterial infections in livestock.
    context: Used primarily in intensive farming operations.
  - use_description: Growth promotion in livestock.
    context: Implemented under specific regulations due to resistance awareness.
side_effects:
  - effect_description: Gastrointestinal upset
    severity: MILD
  - effect_description: Photosensitivity in treated animals
    severity: MODERATE
...
``` |

## System Architecture

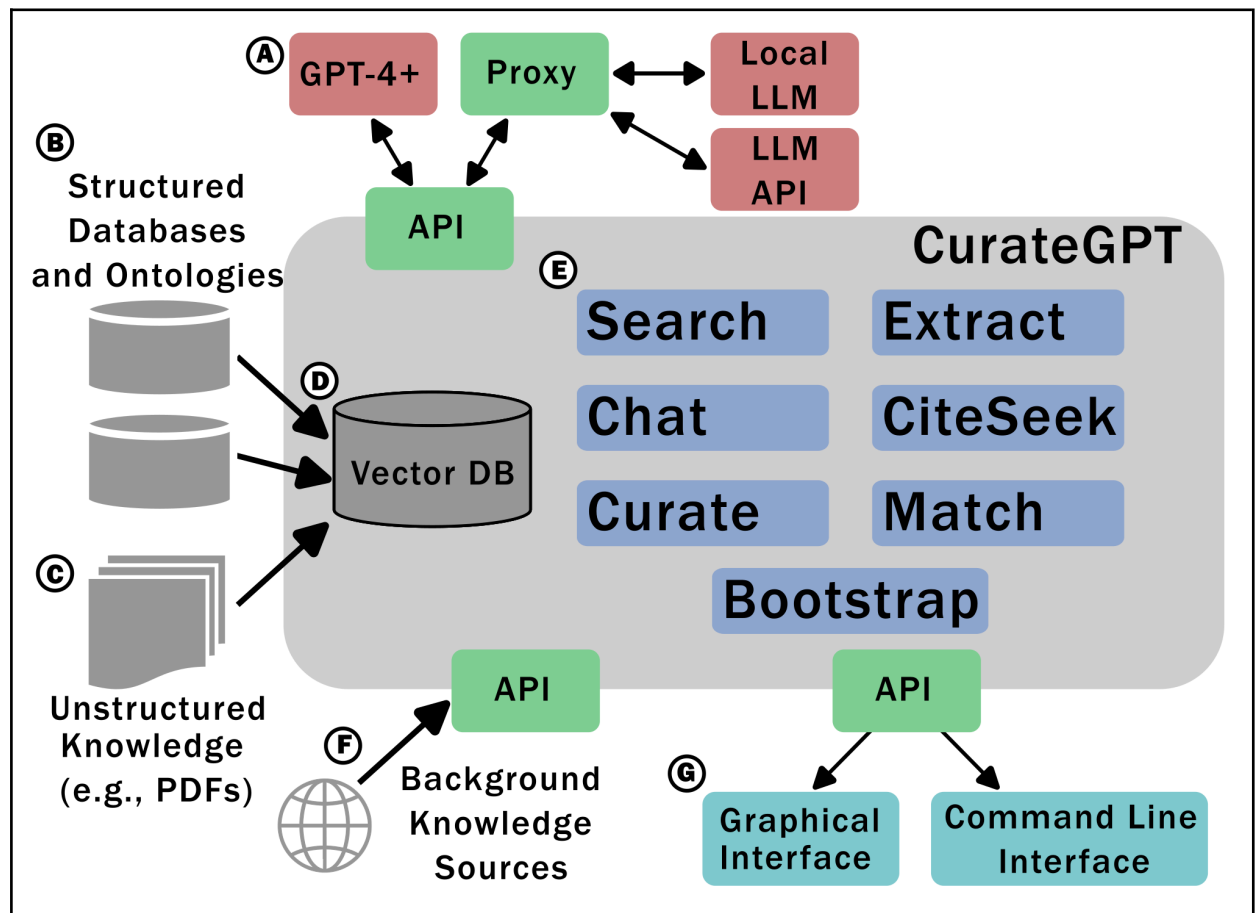

**Figure 1: Overview of the CurateGPT system.** A) The core services leverage pre-trained large language models (LLMs), either through an application programming interface (API) such as the OpenAI API for access to models such as GPT-4 and beyond, or through use of a local proxy (e.g., *litellm*) for access to other LLM APIs or direct execution of locally-available open LLMs such as Llama3. Knowledge sources may be any combination of B) structured databases and ontologies loaded in writable or read-only configurations and C) unstructured text such as that in PDFs. All knowledge sources are indexed in D) a vector database to support retrieval augmented generation (RAG). E) CurateGPT includes functionality driven by tools developed to interface with ontologies to enable semi-automated biocuration supported by known, trusted knowledge sources. These agents and their results may be enriched through relevant text retrieved from background knowledge sources (e.g., PubMed or Wikipedia) and may be accessed by users through G) a graphical interface or a command line interface.

The high-level architecture of CurateGPT is shown in **Figure 1**. In brief, CurateGPT provides direct access to LLM-driven tools for curating structured data from collections of more loosely-structured data and text. The following sections detail how CurateGPT indexes knowledge and supports curation-related tasks.

## Integration with an LLM tooling ecosystem

CurateGPT is part of a thriving and evolving ecosystem of LLMs-powered software tools for supporting data curation tasks. Much of CurateGPT's ability to generate ontology entries (e.g., by building entries based on content from one or more research manuscripts) is built upon our previously described method, Dynamic Retrieval Augmented Generation of Ontologies using AI (DRAGON-AI; (27)). Recognizing the need to associate LLM-extracted information to a consistent set of identifiers, we also constructed the Structured Prompt Interrogation and Recursive Extraction of Semantics (SPIRES) method (18) for performing zero to few-shot information extraction coupled with ontology-based grounding. SPIRES may be used through our OntoGPT software (https://github.com/monarch-initiative/ontogpt) or as part of TALISMAN, a pipeline for gene set summary generation (28). CurateGPT acts as infrastructure for integrating the above methods. We have also found that LLMs are effective tools for querying the structured data relationships within knowledge graphs: our Phenomics Assistant allows biocurators and researchers to use natural language to query a multifaceted biomolecular knowledge graph (29).

The above tools, including CurateGPT itself, do not rely on any specific LLM architecture or infrastructure. They are generalizable to models beyond the popular OpenAI GPT models, including Llama3 and other large open-source models, as well as their corresponding embeddings. These models may be accessed and used with CurateGPT through the *llm* (https://github.com/simonw/llm) and *litellm* (https://www.litellm.ai/) frameworks; the former provides direct access to OpenAI API and a series of plugins for interfacing with other models (e.g., GPT4All models) the latter provides a proxy interface (see **Fig. 1A**) to connect to a variety of local and remote LLM endpoints using the same parameters as the OpenAI API uses.

## Retrieval augmented generation using structured objects

LLMs are highly flexible in their abilities, including complex question-answering, and the capability to shape responses into appropriate data structures. However, without specific tuning or retraining, LLMs will not include knowledge from recently published or domain-specific literature. LLM accuracy may suffer regarding topics the model received limited exposure to during training (30).

A common approach to address these concerns is Retrieval Augmented Generation (RAG) (31). By including the most relevant pieces of information as in-context examples when asking a question of an LLM, RAG can noticeably improve accuracy and relevance of results while reducing the likelihood of LLM hallucination. RAG can be thought of as a kind of dynamic "on-the-fly training" of otherwise fixed models, in which a custom source of documents is indexed in advance, and then queried using semantic similarity to retrieve the most relevant documents for a given query, which are then presented to the model as part of background context.

CurateGPT implements specialized forms of RAG in which combined collections of structured knowledge, unstructured text, and external data may be used to supplement LLM-driven generation and instruction following. We have previously described the technical basis behind these methods in the context of DRAGON-AI, but in brief, this allows either objects or text to be used as queries to retrieve the most relevant objects. Whereas the primary focus of DRAGON-AI is ontology generation, its capability to derive structured knowledge from a variety of data sources is crucial to the functionality of CurateGPT's agents. Where needed, input to an agent is used in semantic similarity search to find the closest matching documents. These documents are then used to provide additional context for generated outputs, enabling inclusion of concepts and details not present in the input query or the LLM's training data.

## Generative-AI-friendly object store

At the core of CurateGPT is a data storage system capable of managing a variety of object types. Here, an "object" is a tree-like structure, equivalent to a JSON object. This allows for flexibility in adapting to different use cases and data, whether structured as tables of measurements, sophisticated nested associations (e.g., the GO-CAM structure (32) used by the Gene Ontology), or documents representing ontology terms. CurateGPT can interface directly with databases as described below. Future work will integrate CurateGPT with the LinkML-Store framework (https://github.com/linkml/linkml-store) for using a broader variety of file systems and data structures.

CurateGPT assists in the generation of meaningful, structured data by searching for relevant text and other objects in background data. In CurateGPT, all objects are stored in a database supporting vector search; the framework currently supports ChromaDB and the DuckDB Vector Similarity Search (VSS) extension. A vector database creates embeddings (i.e., low-dimensional statistical representations) of each object stored within it, allowing for *semantic search* of the store. This is achieved by first creating an embedding of the query (using the same method as is used for the index), and then calculating the nearest *N* objects using vector-based similarity. CurateGPT uses cosine similarity by default but vector search backends support other distance metrics such as Euclidean distance or negative inner product. Queries may be simple terms, complex questions, long narrative text, or even serialized JSON objects. Search results are re-ranked using a Maximal Marginal Relevance (MMR) approach (33). Our goal is to ensure that a diverse set of terms relevant to the search are included, particularly when multiple interrelated ontologies are involved. Further details on this approach are provided in our description of the DRAGON-AI method.

CurateGPT is integrated with a broader data ecosystem. It is capable of retrieving ontologies directly from the Ontology Access Kit (OAK) (https://github.com/INCATools/ontology-access-kit), a feature providing rapid access to current versions of all resources in the OBO Foundry collection (34). To facilitate easy exchange of embeddings, CurateGPT supports a workflow for sharing embeddings and associated metadata. Metadata for embeddings are represented in Vector Embedding Named Object Model indeX (venomx) format, a LinkML-based representation of metadata features such as embedding model, dataset that was embedded, and date of creation (https://github.com/cmungall/venomx). CurateGPT supports a command to upload embeddings and metadata to HuggingFace to enable their sharing and re-use.

## Integrating diverse resources

The CurateGPT backend supports retrieval of structured data from external resources. This includes both loading of local data through an Extract-Transform-Load (ETL) pipeline and retrieval through external APIs. CurateGPT includes a loader capable of loading most ontologies; even large ontologies like ChEBI (35) can easily be ingested in their entirety as part of an initialization pipeline. Ingesting larger resources such as PubMed or Wikipedia in their entirety is more resource intensive (and in this case, may require numerous calls to retrieve embeddings from an LLM) and is instead accomplished through wrappers that call their respective APIs. In these cases, CurateGPT performs relevancy-ranked searches using the resource's public API, then stores the results in its local vector store cache, before performing a more refined embedding-based semantic search. The CurateGPT data retrieval framework includes a novel LLM-powered enhancement: search queries for external APIs are passed to an LLM with a prompt instructing the model to decompose the query into a set of search terms. These search terms are then parsed and fed to the API. See **Table 6** for further examples of resources accessible by CurateGPT and usable with its curation support agents.

**Table 6. External resources accessible through CurateGPT.**

| Wrapper name | Provides access to… |
|---|---|
| *Wrappers supporting dynamic access* | |
| BioProject | Metadata about collections of biological data provided by the NCBI |
| BioSample | Descriptions of biological materials and corresponding experiments, provided by the NCBI |
| ClinVar | Data about clinically-relevant genetic variation, provided by the NCBI |
| Fairsharing.org | Descriptions of data standards, repositories, and journal data policies |
| GitHub issues | Text of issues opened on code and data repositories |
| Google Drive | Text within documents stored on a specified Google Drive |
| JGI Data Portal | Plant, algal, fungal, and microbial genomes and metagenomes curated by the Joint Genome Institute, including those in the Integrated Microbial Genomes (IMG) resource |
| PubMed and PubMed Central | Metadata and text (title and abstract) from PubMed as well as identifiers for PubMed Central full-text documents |
| Wikipedia | Metadata and text of Wikipedia pages |
| *Wrappers supporting data ingest* | |
| BioC NLP results | Textual annotations, including entity and relation extraction results, in BioC format |
| GO-CAMs | Biological relationships structured as Gene Ontology Causal Activity Model objects |
| Google Sheets | Static contents of a Google Sheets spreadsheet |

| HPO annotations | Curated relationships describing associations between Human Phenotype Ontology terms and diseases |
| --- | --- |
| NMDC BioSamples | Descriptions of biological materials and corresponding experiments, provided by the National Microbiome Data Collaborative |
| Ontologies (with Ontology Access Kit) | Local OWL and OBO files, as well as endpoints such as BioPortal, OLS, Ontobee, OntoPortal, and Ubergraph |
| *Wrappers supporting arbitrary data loading* | |
| JSON/YAML | Static loading of JSON or YAML data |
| CSV/TSV | Static loading of comma or tab-delimited data |

# Results

**Figure 2**: **CurateGPT prototype graphical interface.** This annotated screenshot demonstrates the `Curate` agent's functionality in object completion. For this example, we show an example ontology term, but the system is generic and will work with any kinds of annotations or metadata. The left sidebar allows the curator to select (1) the general operation they wish to perform (curate an object); (2)

> which database/collection the object should belong to (here, Ontology for Biomedical Investigations, or OBI); (3) which LLM to use; (4) whether to use SPIRES or basic extraction; (5) an optional source to query for additional context. The main part of the interface allows the curator to enter values for each object field; CurateGPT then automatically completes the rest of the fields. Here the user entered the label (6) for the term they want (magnetoencephalography) and left other fields blank; (7) the user can opt to have the LLM generate a summary which it then uses as background; they click "Suggest" (8) and receive a completed object (9) that shows a generated textual definition (here interleaving both relevant publications as well as similar terms in the database), as well as other relationships. (10) The generated text is parsed into a form in alignment with the desired structure. From here, (11) it may be added to the collection and will be used in subsequent operations (e.g., the user may curate a new term serving as a subclass of magnetoencephalography).

## Example workflow: curation and literature retrieval

Translating observations reported in scientific literature directly into structured, searchable knowledge yields indispensable collections of knowledge, but it remains one of the most laborious tasks performed by biocurators (6,36). CurateGPT can rapidly accelerate this process. For example, a biocurator seeking to add new terms about killifish (a large group of small, freshwater fish, some species of which are used as models of aging and neurodegeneration (37)) to Vertebrate Breed Ontology (VBO) (38) may use the following workflow to find and interpret source literature. To set up the necessary resources, the curator first sets their OpenAI API key, then obtains and indexes a recent version of VBO with the command "make ont_vbo". To save time, this process downloads a pre-built sqlite representation of the ontology.

Once the new database is created, the curator makes it accessible to the CurateGPT interface ("cp -r stagedb/* db/") and opens the Streamlit app ("make app"). The graphical interface is similar to that shown in **Figure 2**.

There is not yet an entry in VBO with the label "killifish", but the curator is curious about whether similar types of fish may be present in the ontology, so they begin with the `Search` command. They set their collection to the newly created *vbo_new* and specify their search string as "killifish". The results indicate no exact results, but general entries for other types of fish are present in VBO (e.g., "Rainbow trout breed"). The curator then uses the *Curate* agent to create a new entry for "Killifish breed" in accordance with the VBO structure and patterns. They enter "Killifish breed" as the label value and do not generate background, though they do include the additional instruction to "Include a definition field with a brief description." (At this point, they may also use another resource such as the NCBI Taxonomy for background, which can help to provide appropriate external identifiers.) The resulting object generated by CurateGPT is shown in **Table 7**.

**Table 7**. Example input and output generated as part of the curation and literature retrieval workflow in CurateGPT. Here, the target data collection is the Vertebrate Breed Ontology and citations are retrieved through the PubMed API.

| Curate | Input | `label: Killifish breed` |
| --- | --- | --- |

|  | Output | `label`: Killifish breed<br>`id`: KillifishBreed<br>`relationships`:<br>- `predicate`: subClassOf<br>  `target`: FishBreed<br>`definition`: A breed of killifish, which are small oviparous fish belonging to the order Cyprinodontiformes. |
|---|---|---|
| Extract | Input | The African turquoise killifish is a breed of killifish. |
|  | Output | `id`: AfricanTurquoiseKillifish_Killifish_<br>`label`: African Turquoise Killifish (Killifish)<br>`definition`: Vertebrate breed of the taxon Nothobranchius furzeri. The African turquoise killifish is recognized for its striking turquoise coloration and rapid aging process, living for only 4-6 months. It's used in scientific research due to its short lifespan, high fecundity, and vertebrate characteristics, making it a valuable model for studying aging and other complex phenotypes.<br>`aliases`: null<br>`relationships`:<br>- `predicate`: subClassOf<br>  `target`: KillifishBreed<br>`logical_definition`: null |
| CiteSeek | Input | `subject`: African Turquoise Killifish<br>`predicate`: is<br>`object`: model for studying aging |
|  | Output | The African Turquoise Killifish (Nothobranchius furzeri) is indeed a model for studying aging. Evidence supporting this assertion can be found in multiple references provided:<br><br>Reference 1 highlights the use of the African Turquoise Killifish as an emerging experimental model in aging research. The killifish is recognized for its short lifespan, making it an ideal subject for studying biological aging processes and providing insights into human aging...<br><br>Reference 1<br>`id`: PMID:26839399<br>`title`: 'The short-lived African turquoise killifish: an emerging experimental model for ageing.'<br>`abstract`: Human ageing is a fundamental biological process …<br>`pmcid`: PMCID:PMC4770150 |

The curator adds this entry to the local CurateGPT collection with the "Add to vbo_new" button. They also wish to create a child entry for a specific breed, the African turquoise killifish. They use the *Extract* function this time, providing a brief text description and additional instructions

("In the description, include the scientific name and at least two sentences describing distinctive features of this breed."). They check the box to generate background information from PubMed entries. The resulting object is shown in **Table 6**.

This entry would likely fit into VBO with few issues. The curator uses the `CiteSeek` agent to find additional references: they select PubMed as the collection to search against this time and provide the full definition from the above object as input. In return, they receive a list of relevant citations as well as an LLM-generated summary of the citations most relevant to the text (**Table 6**). They may then repeat the `Extract` process with the newly generated summary to provide further detail, or they may manually add the citations to the newly curated entry.

## CurateGPT can use design pattern documents to refine suggested terms

Ontologies and knowledge bases frequently include documentation for curators to help guide them in making editorial choices that are consistent with the patterns in the knowledge base as a whole. This documentation might be heavily natural language oriented; or it may be structured according to a computationally-readable design pattern data model, such as is the case for Dead Simple Ontology Design Patterns (DOSDPs) (39). CurateGPT can load either natural language or structured documentation as background knowledge to generate ontology entries better aligned with expectations.

As an experiment, we selected an existing term from the Mondo disease ontology (40) to determine if CurateGPT could generate similar versions of the term when provided or not provided with editor documentation. This term, "bladder urachal squamous cell carcinoma", lacked a definition but otherwise had several relationships within Mondo (e.g., it is a subclass of "bladder squamous cell carcinoma"). Results are shown in **Table 8**. The editor documentation for Mondo includes highly structured DOSDP patterns. We found it was necessary to to pass in additional instructions on how to use the highly technical structures of DOSDPs: "use design pattern documents to ensure definition is populated. Use the "def" pattern for definitions". These instructions may be provided to CurateGPT through the Additional Instructions input in the `Extract` agent, which is analogous to instructing a human curator in how to make use of these patterns. With this additional instruction, CurateGPT was able to generate the missing definition according to the pattern. However, the presence of the design patterns also confused the model and caused it to generate structures that were not consistent with the overall data model loaded in the store. This is likely because the DOSDPs follow an OWL functional style syntax and data model, whereas we loaded in a more OBO-like model for Mondo. It is likely that if the underlying data model used to load Mondo were more OWL-like then these would be more effective. More research is needed here to determine the best ways to use design patterns and editor documentation – this may simply involve providing a textual description of the data model mapping, or using CurateGPT to infer these mappings.

**Table 8. Generation of a Mondo term through the CurateGPT `Extract` agent, with and without Mondo design pattern documentation.** Without any background knowledge, CurateGPT's `Extract` agent is not able to supply a text definition. With the additional background detail provided through structured design patterns (i.e., DOSDPs), and instructions on how to use them, an appropriate definition and aliases are suggested. In the additional instructions, (DOSDP) represents the full text of the YAML document describing patterns used for cancer terms in the ontology; this document is available online at https://github.com/monarch-initiative/mondo/blob/master/src/patterns/dosdp-patterns/cancer.yaml.

|  | Response (no background source) | Response (using DOSDPs) |
| --- | --- | --- |
| Input | `tongue carcinoma` | `tongue carcinoma` |
| Additional instructions | None | Use design pattern documents to ensure definition and annotations are populated. Use the "def" pattern for definitions and the "annotations" to add exact_synonym properties. Design patterns for cancer are as follows: (DOSDP) |
| Output data | `id`: TongueCarcinoma<br>`label`: tongue carcinoma<br>`definition`: A carcinoma that arises from the tongue.<br>`relationships`:<br>- `predicate`: subClassOf<br>  `target`: TongueNeoplasm<br>- `predicate`: subClassOf<br>  `target`: OralCavityCarcinoma<br>`logical_definition`:<br>- `predicate`: subClassOf<br>  `target`: Carcinoma<br>- `predicate`: Disease<br>  `target`: Tongue | `id`: TongueCarcinoma<br>`label`: tongue carcinoma<br>`definition`: A cancer involving a tongue.<br>`aliases`:<br>- malignant tongue neoplasm<br>- malignant neoplasm of tongue<br>- cancer of tongue<br>`relationships`:<br>- `predicate`: subClassOf<br>  `target`: Carcinoma<br>- `predicate`: DiseaseHasLocation<br>  `target`: Tongue<br>`logical_definition`:<br>- `predicate`: subClassOf<br>  `target`: Carcinoma<br>- `predicate`: DiseaseHasLocation<br>  `target`: Tongue |

# Discussion

## Limitations and ongoing challenges

CurateGPT is not designed or intended to replace human curators. Its current agents can supplant some of the more laborious tasks in a curation workflow, such as extracting specific types of information from scholarly texts and identifying sources of supporting details. These tasks are currently supported by a massive array of well-established tools: curators use the search interfaces for structured resources such as PubMed, OMIM, or UniProt to accomplish both of the above tasks and more. CurateGPT's value is in bypassing some of the more frequent questions asked in the course of accomplishing these tasks (e.g., "Are there already entries similar to the one I'm curating in this resource?", or "Am I using the right keywords to search for papers on this subject?") in favor of approaches driven by context and domain knowledge. It also serves to integrate and automate what would otherwise be a collection of repetitive operations. Curators may thereby apply more focus to tasks requiring their deep domain knowledge and curation experience.

We recognize that there is a long way to go in determining how best to evaluate LLM-powered curation support tools and how to unobtrusively integrate them into existing curation workflows. A model's performance on language understanding tasks may not directly translate to its ability to complement a human curator's abilities. Ontologies and other structured knowledge resources vary considerably in their structures and their approaches to data representation. In the absence of documentation such as DOSDPs (see above) the intended scope of any given semantic resource may be unclear to an LLM. Even with these challenges, integrated platforms such as CurateGPT can help to make the production of structured biological and biomedical data more efficient, comprehensive, and practical.

## Ensuring a human-centric curation philosophy

We do not believe that AI can or should replace manual curation. We have a deeply human-centric view of curation, informed by our own professional work and experience in curating knowledge resources, ontologies, and repositories, as well as developing frameworks, interfaces, and algorithms to support these activities. AI can be a powerful assistant, one that can be personalized and interacted with while efficiently supporting the expertise and intuition of curators. The technical capabilities of AI extend well beyond curation support systems that simply present tables of scored predictions. We are encouraged by studies that demonstrate the ability of LLMs to not only enhance productivity, but to help level the playing field in knowledge-centric fields (41,42). Similarly, LLM-driven systems such as GitHub Copilot have seen concrete success in production systems as coding assistants for software developers. These tools offer a crucial feature in that developers can quickly and easily decide whether to accept or reject generated suggestions. Evaluating knowledge is generally more complicated than evaluating code, but both processes may benefit from an effortless interplay between human evaluator and generative process. We have found that making evaluation of generated text as easy as possible is of particular value when combined with RAG approaches, as the final product is a result of at least three sources of knowledge: the human curator, the LLM, and structured external data.

# Acknowledgements

This work was supported by the National Institutes of Health National Human Genome Research Institute [RM1 HG010860]; National Institutes of Health Office of the Director [R24 OD011883]; and the Director, Office of Science, Office of Basic Energy Sciences, of the US Department of Energy [DE-AC0205CH11231 to JHC, JTR, MPJ, HH, NLH, CJM].

# Supplementary Data

**Supplementary Table 1**. Example results of Bootstrap agent used to generate data schema.

```yaml
---
id: https://w3id.org/livestock_antibiotics
name: livestock_antibiotics
description: A knowledge base of antibiotics used in raising livestock
prefixes:
  linkml: https://w3id.org/linkml/
  livestock_antibiotics: https://w3id.org/livestock_antibiotics
imports:
  - linkml:types
classes:
  Container:
    tree_root: true
    attributes:
      members:
        range: Antibiotic
        multivalued: true
        inlined_as_list: true
  Antibiotic:
    description: A representation of an antibiotic used in livestock
    attributes:
      name:
        required: true
        identifier: true
      description:
        range: string
      associated_species:
        range: Species
        multivalued: true
        inlined_as_list: true
      common_uses:
        range: UseCase
        multivalued: true
        inlined_as_list: true
      side_effects:
        range: SideEffect
        multivalued: true
        inlined_as_list: true
  Species:
    description: A species that may be treated with the antibiotic
    attributes:
      scientific_name:
        required: true
        identifier: true
      common_name:
        range: string
      classification:
        range: string
  UseCase:
    description: Use cases for the antibiotic in livestock
```

```yaml
    attributes:
      use_description:
        required: true
        identifier: true
      dosage:
        range: string
  SideEffect:
    description: Possible side effects from using the antibiotic
    attributes:
      effect_description:
        required: true
        identifier: true
      severity:
        range: SeverityLevelEnum
enums:
  SeverityLevelEnum:
    permissible_values:
      LOW: {}
      MEDIUM: {}
      HIGH: {}
```

**Supplementary Table 2**. Example results of Bootstrap agent used to generate data.

```yaml
---
name: Tetracycline
description: A broad-spectrum antibiotic used in various livestock species.
associated_species:
  - scientific_name: Bos taurus
    common_name: Cattle
    classification: Mammalia
  - scientific_name: Gallus gallus domesticus
    common_name: Chicken
    classification: Aves
common_uses:
  - use_description: Treatment of bacterial infections such as respiratory illness
    dosage: 10 mg/kg body weight, orally once a day
  - use_description: Control of chronic respiratory disease
    dosage: 20 mg/kg body weight, orally once a day for one week
side_effects:
  - effect_description: Gastrointestinal disturbances
    severity: MEDIUM
  - effect_description: Hepatotoxicity in rare cases
    severity: HIGH
---
name: Penicillin
description: A widely used antibiotic primarily for treating infections in livestock.
associated_species:
  - scientific_name: Sus scrofa domesticus
    common_name: Pig
    classification: Mammalia
  - scientific_name: Ovis aries
    common_name: Sheep
    classification: Mammalia
```

```yaml
common_uses:
  - use_description: Treatment of bacterial pneumonia
    dosage: 5 mg/kg body weight, intramuscularly, twice a day
  - use_description: Treatment of bacterial mastitis
    dosage: 8 mg/kg body weight, intramuscularly, once a day
side_effects:
  - effect_description: Allergic reactions
    severity: HIGH
  - effect_description: Nephrotoxic effects in prolonged usage
    severity: MEDIUM
---
name: Florfenicol
description: An antibiotic effective against a wide range of bacteria.
associated_species:
  - scientific_name: Oncorhynchus mykiss
    common_name: Rainbow trout
    classification: Actinopterygii
  - scientific_name: Anas platyrhynchos domesticus
    common_name: Duck
    classification: Aves
common_uses:
  - use_description: Treatment of skin infections in fish
    dosage: Add 10 mg/L in water
  - use_description: Control of bacterial infections in poultry
    dosage: 30 mg/kg body weight, orally once a day
side_effects:
  - effect_description: Reduced feed intake
    severity: LOW
  - effect_description: Potential bone marrow suppression
    severity: HIGH
```